\documentclass[10pt, conference, compsocconf]{IEEEtran}

\usepackage[normalem]{ulem}

\usepackage{amsmath}
\usepackage{url} 
\usepackage{xspace} 
\usepackage{graphicx} 
\usepackage{comment} 
\usepackage{ulem}
\usepackage{subfig}
\usepackage{tabu}
\usepackage[table]{xcolor}
\usepackage{caption}
\usepackage{array,multirow}
\usepackage[]{algorithm2e}
\usepackage{amssymb}
\usepackage{hyperref}

\let\oldemptyset\emptyset

\newcolumntype{P}[1]{>{\centering\arraybackslash}p{#1}}
\newcolumntype{M}[1]{>{\centering\arraybackslash}m{#1}}

\definecolor{tableShade}{gray}{0.95} 
\begin{document}
%
% paper title
% can use linebreaks \\ within to get better formatting as desired
\title{Algorithmic Performance-Accuracy Trade-off\\ in 3D Vision Applications Using HyperMapper}

%Algorithmic Parameter Search in 3D Vision Applications Using }

% author names and affiliations
% use a multiple column layout for up to two different
% affiliations

\author{\IEEEauthorblockN{Luigi Nardi\IEEEauthorrefmark{1}, Bruno                   
Bodin\IEEEauthorrefmark{2}, Sajad Saeedi\IEEEauthorrefmark{1}, 
%Harry Wagstaff\IEEEauthorrefmark{2},
} \\                 
\IEEEauthorblockN{
Emanuele             
Vespa\IEEEauthorrefmark{1}, 
%Bj\"{o}rn Franke\IEEEauthorrefmark{2}, 
Andrew J. Davison\IEEEauthorrefmark{1}, 
%Mike O'’Boyle\IEEEauthorrefmark{2}, 
Paul H. J. Kelly\IEEEauthorrefmark{1}} \\             
\IEEEauthorblockA{\IEEEauthorrefmark{1}Department of Computing,                     
Imperial College London\\                                                           
London, UK\\                                                                        
\{l.nardi, s.saeedi, e.vespa14, ajd, p.kelly\}@imperial.ac.uk}         
\IEEEauthorblockA{\IEEEauthorrefmark{2}Institute for Computing Systems Architecture,
The University of Edinburgh\\                                                       
Edinburgh, Scotland\\                                                               
%\{bbodin
%, h.wagstaff, bfranke, mob
%\}@inf.ed.ac.uk}                            
bbodin@inf.ed.ac.uk}
}

% make the title area
\maketitle

\begin{abstract}
%Other names for the framework: 
% Design Hyperspace Autopilot
% Design Hyperspace Autopilot Research Methodology Artifact (DHARMA).
% Generalised Automatic Understanding of Design Interactions (GAUDI).
% Blithe
% HyperMapper
% ParetoDaemon
% Randm Forest Active Learning (RaFALe)

%Automatic search of optimal configurations  have been proven to be highly effective for wide range of computer vision applications. The exploration goal is to find optimal configurations that maximise given quality of result goals.
In this paper we investigate an emerging application, 3D scene understanding, likely to be significant in the mobile space in the near future.  The goal of this exploration is to reduce execution time while meeting our quality of result objectives. In previous work, we showed for the first time that it is possible to map this application to power constrained embedded systems, highlighting that decision choices made at the algorithmic design-level have the most significant impact. \\
\indent As the algorithmic design space is too large to be exhaustively evaluated, we use a previously introduced multi-objective random forest active learning prediction framework dubbed HyperMapper, to find good algorithmic designs. We show that HyperMapper generalizes on a recent cutting edge 3D scene understanding algorithm and on a modern GPU-based computer architecture. HyperMapper is able to beat an expert human hand-tuning the algorithmic parameters of the class of computer vision applications taken under consideration in this paper automatically. 
In addition, we use crowd-sourcing using a 3D scene understanding Android app to show that the Pareto front obtained on an embedded system can be used to accelerate the same application on all the 83 smart-phones and tablets with speedups ranging from 2x to over 12x. 
\end{abstract}

\begin{IEEEkeywords}
design space exploration; machine learning; computer vision; SLAM; embedded systems; GPU; crowd-sourcing;
\end{IEEEkeywords}

%But you might think about mapping Pareto-optimal configurations to support dynamic adaptation, automatically selecting the best combination of algorithmic parameters for a given scene and power-performance objective.

%The point is that a human might pick one good design.  We can generate a whole Pareto front, with hundreds of design alternatives each optimal for a given situation – so we can dynamically adapt while being sure to be close to optimal whatever our context.  This requires far more work than a human would/could do.  The human gets to focus on *setting* the objectives.

% For peer review papers, you can put extra information on the cover
% page as needed:
% \ifCLASSOPTIONpeerreview
% \begin{center} \bfseries EDICS Category: 3-BBND \end{center}
% \fi
%
% For peerreview papers, this IEEEtran command inserts a page break and
% creates the second title. It will be ignored for other modes.
\IEEEpeerreviewmaketitle

\section{Introduction} 
\label{section:introduction}

Motivated by the increasing complexity of hardware
and software components, automatic performance tuning
techniques have flourished in the past few years. While
closed-form mathematical performance models have been
successfully applied to compiler optimizations, they often lack of accuracy or expressive
power, which could undermine the ability to capture the
complex interactions that occur between tool-chain and
hardware parameters. This intricacy is exacerbated when
including algorithmic parameters in the tuning practice,
where deep domain knowledge may be required to meet
multiple conflicting design goals.
This paper extends the work done in \cite{Bodin_PACT_2016} where the HyperMapper framework was introduced. The authors showed how going beyond conventional benchmarking in computer systems research is possible by exposing the algorithmic-level design space. They used a vertical approach to design and program heterogeneous MPSoCs exploring all levels of the stack, from compilers to the micro-architecture, to optimally map the executed code onto such diverse hardware resources. The authors found that the algorithmic space enables important approximate computing research exploring trade-off in accuracy and performance.  
The rationale behind including the algorithmic parameters in the design space exploration is that although these algorithms are tuned for desktop systems, the same configurations are not optimal in a mobile MPSoC setting.

%Traditionally, system designers have evaluated new architecture and compiler features using well-studied and broadly accepted benchmarks such as SPEC2006~\cite{2006Spec}. 
%However, since such benchmarks represent a historical snapshot of applications, they are not representative of modern requirements.  
%In contrast, to design tomorrow's systems, we need to consider new emerging applications from diverse domains.

% \begin{figure*}
% 	\centering
% 	\includegraphics[height=1.2in]{./img/kfusion_coarse.pdf}
% 	\caption[KFusion taskgraph]{Key computational steps of the KFusion algorithm represented as a task graph. 
% 		Each task comprises one or more OpenCL kernels as depicted.
%         } 
% 	\label{fig:kfusion}
% \end{figure*}

In this paper we focus on one set of emerging applications that is becoming significant in the mobile space: real-time 3D scene understanding in computer vision.   
In particular, we investigate dense simultaneous localization and mapping (dense SLAM) algorithms which are extremely computationally demanding. 
One such dense SLAM algorithm is KinectFusion~\cite{2011Newcombe} (KFusion) which estimates the pose of a depth camera whilst constructing a highly detailed 3D model of the environment. 
Another well-known dense SLAM algorithm is the ElasticFusion algorithm \cite{Whelan_RSS_2015}. While in KinectFusion the map is shown by dense voxels, in ElasticFusion, the map is shown by small disc-shaped objects called surfels. Unlike KinectFusion, ElasticFusion has loop closure functionality built in and uses both the depth and the RGB cameras. In this paper, KinectFusion and ElasticFusion are the benchmarks on which the experiments are performed. Since these algorithms are typically tuned for high-end desktops with high power budget, executing them on power-constrained embedded devices is very challenging and, therefore, represents a realistic future application use case.  
We use the SLAMBench benchmarking framework~\cite{2015PAMELASLAMBench}, 
which contains KFusion \cite{2011Newcombe} and ElasticFusion \cite{Whelan_RSS_2015} implementations, as it allows us to capture the performance metrics used to drive our design space exploration.  

We define the performance in terms of accuracy of estimated trajectory (in centimeters, lower is better) and runtime (measured as wall clock time per frame in seconds, lower is better). 
The runtime is sometimes also quantified by the number of frames processed in one second, i.e. frames per second (FPS), higher is better; 
the current Microsoft Kinect (or equivalent ASUS Xtion Pro) RGB-D sensor runs at 30 FPS, so 30 FPS is needed for real-time processing. 
These two metrics interact and are considered simultaneously for a through  evaluation of the system. 

Since the algorithmic design space can be extremely large, it is not feasible to try all possible configurations. 
Instead, we sample the domain space and automatically build a model that predicts the two performance metrics for a given configuration.
Using this model, and a methodology from machine learning known as active learning, we predict a two dimensional performance Pareto-optimal configurations curve that can be then stored on the machine to support dynamic adaptation, automatically selecting the best combination of algorithmic parameters for a given scene and accuracy-performance objective. While a human might pick one good design, we generate a whole Pareto front, with hundreds of design alternatives each optimal for a given situation. This enable us to dynamically adapt while being sure to be close to optimal whatever our context.  This requires far more work than a human could do. The human gets to focus on setting the objectives.

In previous work \cite{Bodin_PACT_2016}, by exploring the resulting Pareto curve on the KFusion application we obtain a mapping to an embedded platform that results in a 6.6-fold speedup over the original mobile implementation.  More precisely, this new configuration runs at nearly 40 FPS while maintaining an acceptable accuracy (under 5 cm localization error) 
and keeping power consumption under 2 Watts. The Pareto front contains many more configurations, allowing us to trade between runtime, power consumption, and accuracy, depending on our desired goals. For example, we can also find points which minimize power consumption (e.g., a configuration providing 11.92 FPS at 0.65W) or optimized for execution time without exceeding a given power budget (29.09 FPS at less than 1W). 

Additionally, in a recent work \cite{Saeedi_ICRA_2017}, we have demonstrated that the design space exploration can be extended to include physical parameters such as the motion of the camera and the structure of the environment. By this extension, not only the performance of the SLAM algorithm is improved, but also the robustness is increased.

This paper demonstrates that HyperMapper generalizes on a recent cutting edge 3D scene understanding algorithm, i.e. ElasticFusion \cite{Whelan_RSS_2015}. 
By exploring the resulting Pareto curve we obtain a mapping to a modern discrete GPU-based system which is a very similar machine to the one used by the ElasticFusion developers. That results in a 1.52-fold speedup over the original design defined by the default configuration while also improving accuracy. The default configuration was defined by the the original developers of ElasticFusion. Another configuration shows a 2-fold improvement in accuracy (2.69 cm) compared to the default configuration (5.58 cm) with a speedup of 1.25.
HyperMapper is able to beat an expert human hand-tuning the algorithmic parameters of the class of Computer Vision applications taken under consideration in this paper automatically. 

In addition, we use crowd-sourcing using the Android SLAMBench KFusion app to show that the Pareto front obtained on an embedded system can be used to speed up the same application on all the 83 smart-phones and tablets crowd-sourced with speedups ranging from 2 to over 12. 

The main contributions of the paper are:
\begin{itemize} 
\item We demonstrate how HyperMapper's algorithmic design-space exploration generalizes across two very different applications, i.e. KFusion and ElasticFusion, and on multiple devices. 
\item In order to explore the potential for this approach we evaluate our methodology on an emerging SLAM benchmarking framework, i.e. SLAMBench, which supports quantitative evaluation of solution accuracy and execution time. 
On the new application considered, ElasticFusion, we obtain a a 1.52x best improvement in execution time and 2-fold improvement in accuracy over an hand-tuned implementation by a SLAM domain expert. 
\item We show how the algorithmic Pareto front learned on one device speeds up a variety of smart-phones and tablets evaluated using a crowd-sourcing experiment. 
\end{itemize}

%##########################################2 MOTIVATION & BACKGROUND ############################
\section{Background}
\label{section:motivation}

Simultaneous localization and mapping (SLAM) systems aim to perform real-time localization and mapping ``simultaneously'' from a sensor moving through an unknown environment. 
Localization typically estimates the location and pose of the sensor with respect to a map which is extended as the sensor explores the environment.
Dense SLAM systems in particular map entire 3D surfaces, as opposed to non-dense (feature-based) systems where maps are represented at the level of sparse point landmarks.
Dense SLAM systems enable a mobile robot to perform path planning and collision avoidance, 
or an augmented reality (AR) system to render physically plausible animations at appropriate locations in the scene.
%~\cite{2014ISMAR,2015ICHR}. 
Recent advances in computer vision have led to the development of real-time algorithms for dense SLAM such as KFusion~\cite{2011Newcombe} and ElasticFusion \cite{Whelan_RSS_2015}. Such algorithms estimate the pose of a depth camera while building a highly detailed 3D model of the environment (see \cite{SLAMBench}). 

Such real-time 3D scene understanding capabilities can radically change the way robots interact with the world.
%~\cite{2015ICHR}.  
While classical feature-based SLAM techniques are now crossing into mainstream products via embedded implementations, such as Project Tango~\cite{ProjectTango} and Dyson 360 Eye~\cite{DysonLab}, \textit{dense} SLAM algorithms with their high computational requirements 
are largely at the prototype stage on GPU-based PC or laptop platforms~\cite{2011Newcombe,Whelan_RSS_2015}. 
However, when running in an embedded context, it is not feasible to include a large GPU with high power and cooling requirements. 
In addition, even when running on a desktop system it is important to run on an optimal design configuration because this would save machine resources that would allow to reduce the overall system latency. 
While offloading to a remote machine is possible in some circumstances, this can introduce additional latency which makes it unsuitable for real-time situations such as AR or UAV navigation applications.

KFusion registers and fuses the stream of measured noisy depth frames from a depth camera (such as Microsoft Kinect), as the scene is viewed from different viewpoints, into a clean 3D geometric map. 
It is beyond the scope of this paper to go into the details of the KFusion algorithm, we briefly outline the key computational steps involved, for more information the reader should refer to \cite{Bodin_PACT_2016}. SLAMBench provides multiple implementations of the KFusion algorithm. We use the OpenCL implementation, and execute each OpenCL kernel on the GPU of our target platforms.

KFusion normalizes each incoming depth frame and applies a bilateral filter (\textit{Preprocessing}) to reduce noise. 
In the \textit{Tracking} step, it computes a point cloud (with normals) for each pixel in the camera frame of reference and 
estimates the new 3D pose of the moving camera by registering this point cloud with the current global map using iterative closest point (ICP)~\cite{1992Besl}. 
Once the new camera pose has been estimated, the corresponding depth map is fused into the current 3D reconstruction (\textit{Integration}).  
KFusion utilizes a voxel grid as the data structure to represent the map, employing a truncated signed distance function (TSDF) to represent 3D surfaces. 
The 3D surfaces are present at the zero crossings of the TSDF and can be recovered by a \textit{Raycasting} step, which is also useful for visualizing the reconstruction.

ElasticFusion is an incremental, dense SLAM algorithm, which supports local and global loop closure without employing an explicit pose graph. The algorithm  creates  a  surfel-based  model  of  the  environment. Each  frame attempts to perform local loop closures - to stay close to the mode of the map distribution, and also global loop closures - to avoid drift and maintain global consistency.

In this work we use the SLAMBench benchmarking framework~\cite{2015PAMELASLAMBench} which enables evaluation of runtime and accuracy for KFusion and ElasticFusion.
SLAMBench is provided with an absolute trajectory error (ATE) metric. 
The ATE is domain-specific accuracy metric calculated as the mean difference between the real trajectory and the estimated trajectory of a camera produced by a SLAM implementation. 
Thus, smaller ATE implies less deviation from the real trajectory.

\begin{figure}[tb] 
\includegraphics[width=1.0\columnwidth, trim={4.0cm 3.0cm 4.5cm 3.0cm}, clip]{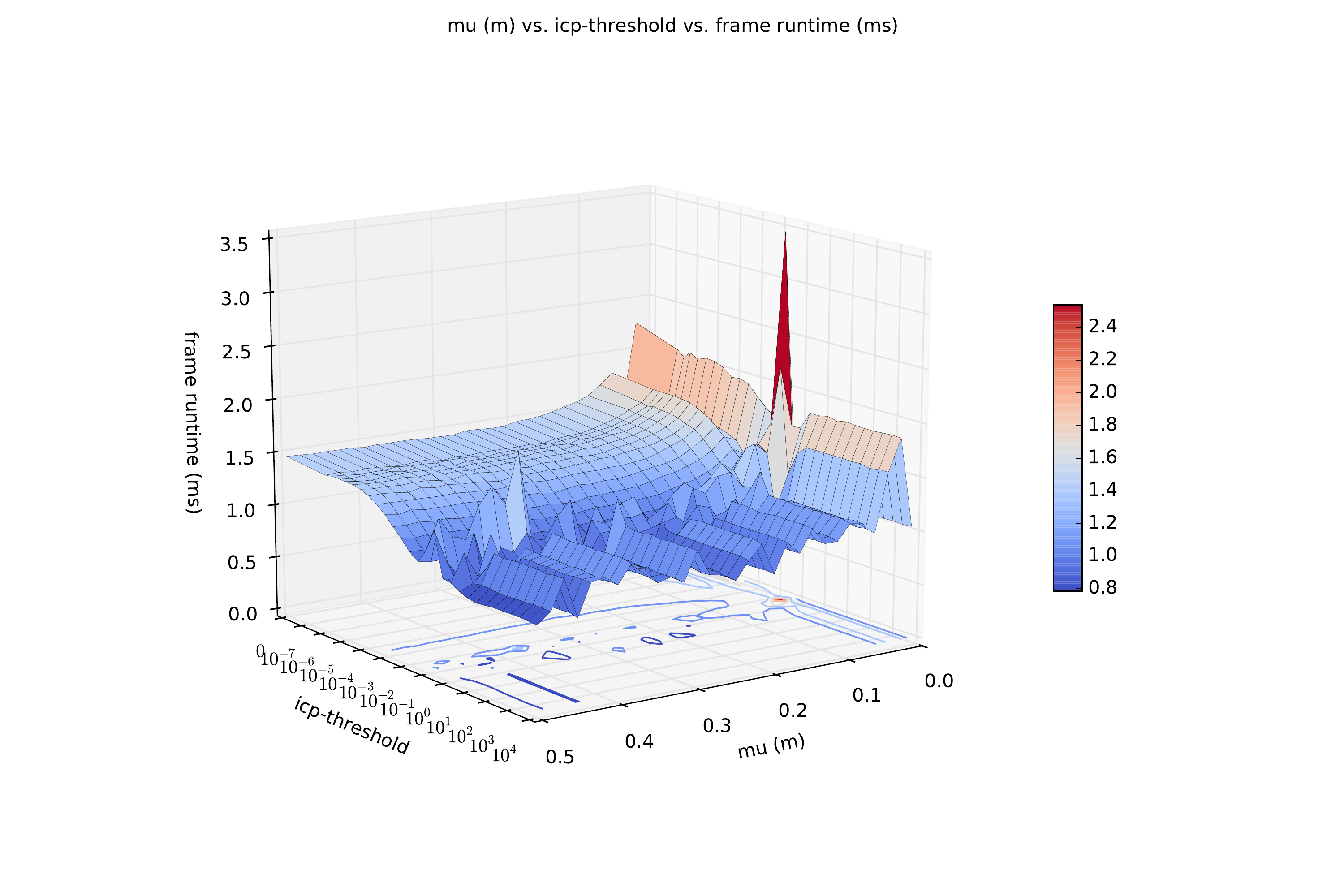} 
\caption{KFusion runtime response surface when varying just two parameters $mu$ and $icp-threshold$, keeping the rest of the parameters as for the default configuration. 
This shows the non-convexity, multi-modality and non-smoothness between configurations.
%\fixmeS{labels are not legible, how about using psfrag package?}
}
\label{figure_non_convexity} 
\end{figure}

Hand optimization of algorithmic parameters in SLAM applications is in general not feasible. Fig. \ref{figure_non_convexity} shows the KFusion runtime response surface when varying two algorithmic parameters, keeping the rest of the parameters as for the default configuration. The picture depicts a non-convex, multi-modal and non-smooth runtime response surface which is in general very difficult to hand-tune by trial and error.

%########################################  Methodology ###########################
\section{Design Spaces and Methodology}
\label{section:methodology}

In this section we describe our approach, including a detailed explanation of the design space parameters and the objectives which we are targeting. In Section \ref{section:machine learning}, we go on to describe the search techniques we use to guide our exploration through the design space. The methodology used is the same as the one in \cite{Bodin_PACT_2016}, the reader can refer to that work for more information. 
% \fixmeL{It seems that Google Scholar didn't take it into account some of our citations: to the PACT paper and to the Diplomat paper} 

\subsection{Experimental Setting}

In order to evaluate our design space exploration (DSE) we use the SLAMBench framework with the ICL-NUIM \cite{2014Handa,Handa:etal:arXiv2015} dataset, specifically the first 400 frames of living room  trajectory 2. 
We halved the original sequence in order to reduce the overall execution time of the benchmark; this was done after careful consideration that the accuracy metric is still representative of the whole sequence.

Usual approaches in performance optimization consider benchmark suites that are, in general, 
a set of small kernels extracted from real applications. 
A criticism to what can be learnt from a benchmark suite is that they may not well represent and 
capture the complex interaction of kernels in a real-world application. 
Our applications are composed of more than 10 GPU-accelerated kernels. 
They present the opportunity to tackle  exploration of parameters at the algorithmic level that is not possible with conventional benchmark suites.

During execution, the following two performance metrics are collected: 
1) computation time and, 
2) absolute trajectory error (ATE) of the frame sequence.

\subsection{KFusion Design Space}
\label{section:methodology:KFusionDSE}

%The possible values of the parameters taken into consideration for the design space exploration of the algorithmic parameters are summarized in Table \ref{tabParameters} for the KFusion and ElasticFusion.

We summarize the algorithmic parameters that mostly affect our performance metrics. In the case of the SLAMBench implementation of the KFusion algorithm, we have access to the listed parameters.
An extensive explanation of these can be found in \cite{2011Newcombe, 2015PAMELASLAMBench}. 

\begin{itemize}
\item \begin{bf}Volume resolution\end{bf}: 
The resolution of the scene being reconstructed. 
As an example, a 64x64x64 voxel grid captures less detail than a 256x256x256 voxel grid. 
\item \begin{bf}$\boldmath\mu$ distance\end{bf}: 
The output volume of KFusion is defined as a truncated signed distance function (TSDF) \cite{2011Newcombe}. Every volume element (voxel) of the volume contains the best likelihood distance to the nearest visible surface, up to a truncation distance denoted by the parameter $\mu$, also referred as $mu$ in the text.
\item \begin{bf}Pyramid level iterations\end{bf}: The number of block averaging iterations to perform while building each level of the image pyramid.
\item \begin{bf}Compute size ratio\end{bf}: 
The fractional depth image resolution used as input. As an example, a value of 8 means that the raw frame is resized to one-eighth resolution. 
\item \begin{bf}Tracking rate\end{bf}: The rate at which the KFusion algorithm attempts to perform localization. A new localization is performed after every tracking rate number of frames.
\item \begin{bf}ICP threshold\end{bf}: The threshold for the iterative closest point (ICP) algorithm \cite{1992Besl} used during the tracking phase. 
\item \begin{bf}Integration rate\end{bf}: 
As the output of KFusion is a volumetric representation of the recorded scene, it needs to be repeatedly expanded using new frames. 
A new frame is integrated after every integration rate number of frames.
\end{itemize}
We observe that the KFusion algorithmic design space consists of roughly 1,800,000 points. 

%Furthermore, the exploration of algorithmic parameters involves trade-offs between accuracy and runtime. 

\subsection{ElasticFusion Design Space}
\label{section:methodology:ElasticFusionDSE}

We summarize the algorithmic parameters considered in the ElasticFusion design exploration via the SLAMBench framework. An extensive explanation of these can be found in \cite{Whelan_RSS_2015}. 

%\fixmeL{Is it a good idea to remove the table on the KFusion and ElasticFusion parameters? That just makes things clearer for the reader.}
In general, there are two categories of parameters in ElasticFusion: algorithmic parameters, thresholds and flags. The algorithmic parameters and thresholds considered in our exploration are:
\begin{itemize}
\item \begin{bf}ICP/RGB weight\end{bf}: 
Relative ICP/RGB tracking weight. Incremental pose estimation is done both in the photometric RGB space and the geometric depth space. Then the results are merged by applying this weight parameter.
\item \begin{bf}Depth cut off\end{bf}: 
Cutoff distance for depth processing. The algorithm ignores the raw depth input larger than this threshold. 
\item \begin{bf}Confidence threshold\end{bf}: Surfel confidence threshold. As a surfel is observed more, the confidence of the surfel increases. Once the confidence of the surfel is larger than a threshold, it is included in the processing pipeline. Lowering this threshold will create a noisy map. 
%\item \begin{bf}ICP error threshold\end{bf}: Local loop closure residual threshold.
%\item \begin{bf}ICP count threshold\end{bf}: Local loop closure inlier threshold.
%\item \begin{bf}ICP error threshold\end{bf} and \begin{bf}ICP count threshold\end{bf}: Both these thresholds are used in the local loop closure.
%\item \begin{bf}Covariance threshold\end{bf}: Local loop closure covariance threshold. When matching two consecutive frames, a covariance matrix is generated. A large convarince matrix means that the tracking is lost.
%\item \begin{bf}Photometric threshold\end{bf}: Global loop closure photometric threshold. A fern database includes all keyframes used for the global loop closure. When matching to a frame in the fern database for a global loop closure, the matched frame photometric error must be lower than this threshold value.
%\item \begin{bf}Fern threshold\end{bf}: Fern encoding threshold. This threshold decides whether or not to add a frame to the fern database.
\end{itemize}
The flags are:
\begin{itemize}
%\item \begin{bf}Confidence threshold\end{bf}: Surfel confidence threshold.
%\item \begin{bf}ICP error threshold\end{bf}: Local loop closure residual threshold.
%\item \begin{bf}ICP count threshold\end{bf}: Local loop closure inlier threshold.
%\item \begin{bf}Covariance threshold\end{bf}: Local loop closure covariance threshold.
%\item \begin{bf}Photometric threshold\end{bf}: Global loop closure photometric threshold.
%\item \begin{bf}Fern threshold\end{bf}: Fern encoding threshold.
\item \begin{bf}Disable SO3 pre-alignment\end{bf}: While tracking, setting this flag disables pre-alignment in 3D rotation group, known as SO(3).
\item \begin{bf}Open loop\end{bf}: Setting flag disables the local loop
closure code in ElasticFusion,
\item \begin{bf}Relocalisation\end{bf}: By setting this flag, ElasticFusion attempts to relocate its pose, i.e.`get back on track' if it is lost,
% \item \begin{bf}Pyramid\end{bf}: By setting this flag, a pyramid of images is created and used for tracking in the RGB odometry.
\item \begin{bf}Fast odometry\end{bf}: By setting this flag, the RGB odometry uses only a single level pyramid, hence faster processing.
\item \begin{bf}Frame to frame RGB\end{bf}: Setting this flag enables frame-to-frame RGB tracking.
\end{itemize}
We observe that the ElasticFusion algorithmic design space consists of roughly 450,000 possible configurations. 
Exploration on ElasticFusion is a work in progress, additional parameters that significantly affect performance will be considered in future explorations, e.g. compute size ratio, ICP error threshold, ICP count threshold, covariance threshold, photometric threshold, and fern threshold.

\subsection{Multi-Objective Optimization Goal}
%Figure \ref{fig:prediction} presents a fictitious example depicting samples (in green) over a 2-dimensional optimization space 
% \begin{figure}
% \includegraphics[width=1.0\columnwidthMotivated by the increasing complexity of hardware
% \caption[Prediction Chart]{
% Illustrative example based on fictitious data.  
% This is a two-objective optimization goal in the error and runtime performance metrics. 
% The samples in green are spread all over the space. 
% We are interested in the region highlighted by the black circle, namely the targeted prediction area.
% The Pareto front is represented in blue. 
% }
% \label{fig:prediction}
% \end{figure}
% \footnote{For visualization purposes we are only showing two performance metrics, 
% namely the error and the runtime.}. 
% In order to meet the runtime and accuracy thresholds (in dashed lines), the solutions 
% of our exploration are confined to the bottom left region of the space, the targeted prediction area (in black). 
In a multi-objective optimization setting, a single solution that minimizes all performance metrics simultaneously does not exist in general. 
Therefore, attention is paid to Pareto-optimal solutions --- that is, solutions that cannot be improved in any of the objectives without degrading at least one of the other objectives. 

\subsection{Design Space Exploration Tool}
\label{section:machine learning}
The algorithmic parameter space we are investigating is too large to be exhaustively evaluated on the hardware platform.  
We use HyperMapper introduced in \cite{Bodin_PACT_2016} to a cheaper route
of training a predictive machine learning model over a handful of examples (points in the parameter space) evaluated on hardware. 
HyperMapper accurately create a surrogate model and predicts the performance over the entire parameter space, while being many orders of magnitude faster as compared to running the application on hardware over a video sequence for big parameter settings. Unfortunately since we 
do not know the performance over the parameter space, we are also unaware of the points for which running a physical experiment will be 
most informative, in the sense of yielding the greatest increase in the prediction accuracy of our model - a classic chicken and egg problem.
 Thus, we resort to bootstrapping predictive models (two separate randomized decision forests~\cite{breiman84book} for accuracy and runtime prediction) from a small number of randomly drawn samples
 in the parameter space. These models are then refined in subsequent iterations by drawing more samples from the parameter space (and retraining over the collective set); 
 the new samples are now drawn to implicitly maximize the prediction accuracy near the respective Pareto optimal fronts. This strategy of letting the predictive model decide which samples 
will be most beneficial in increasing predictive accuracy over unseen regions of the parameter space is called active %learning~\cite{cohn94ml,warmuth01nips}.
learning~\cite{cohn94ml}.
% Note that we explored a number of base predictive models including artificial neural networks, support vector machines, and nearest neighbors.
% Our experiments indicated that randomized decision forests outperform these methods, thus we stick to this class of models throughout this paper. 

\begin{figure}[tb]
\includegraphics[width=1.0\columnwidth]{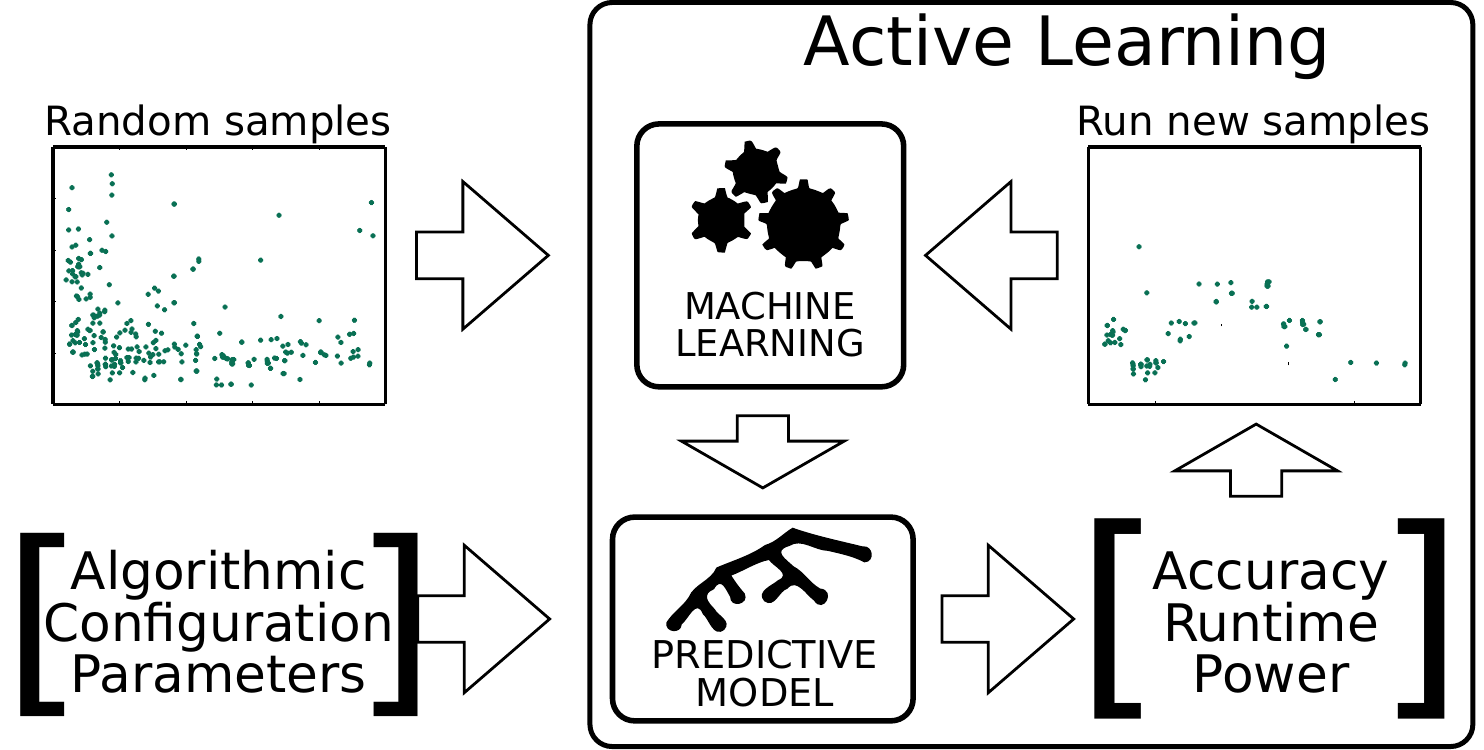}
\caption[Learning Chart]{
The learning step is based on a tiny subset of the overall algorithmic space; these are the samples that are actually run. 
Subsequently, the predictive model can predict accuracy and performance of an unseen configuration depending on its parameters.}
\label{fig:ML}
\end{figure}

The combination of many weak regressors (binary decisions) allows approximating highly non-linear and multi-modal functions with great accuracy.
HyperMapper trains separate regressors to learn the mapping from our input (parameter) space to each output variable, i.e. the two performance metrics. 
This methodology is depicted in Figure \ref{fig:ML} from the original work \cite{Bodin_PACT_2016} and explained in the next sections. Refer to the original paper for more information.

Active learning is a paradigm in supervised machine learning which uses fewer training examples to achieve better 
prediction accuracy - by iteratively training a predictor, and using the predictor in each iteration to 
choose the training examples which will improve its performance over a predefined objective. 
%increase its accuracy the most. 
Thus the accuracy of the predictive model is incrementally improved by interleaving exploration and exploitation steps, 
as shown by the feedback loop in Figure~\ref{fig:ML}. 
% We initialize our base predictors (randomized decision forests) from a very small number of randomly sampled points in the parameter space.
% For these points the application is evaluated over a video sequence on the hardware platform, yielding accuracy, runtime, and power consumption corresponding to these points (labels in a supervised setting). 
Since our objective 
is to accurately estimate the points near the Pareto front, we use the current predictor to provide performance values 
over the entire parameter space and thus estimate the Pareto fronts for accuracy and runtime (separately). For the next iteration, 
only parameter points near the predicted Pareto front are sampled (and evaluated on hardware), and subsequently used to train 
new predictors using the entire collection of training points from current and all previous iterations. This process is repeated over 
a number of iterations. 
% Our experiments (Sect. \ref{evaluation}) indicate that this smarter way of searching for 
% highly informative parameter points in fact yields superior predictors as compared to a baseline that uses randomly sampled 
% points alone. 
% Thus by iterating this process several times in the active learning loop, we are able to discover high-quality design configurations that lead to good performance outcomes. 
% This approach enables predicting performance metrics over a parameter space comprising of approximately 1,800,000 points on KFusion in a matter of seconds and similarly for ElasticFusion.

% \fixmeL{When we prepare the camera ready we should add a variable for the max number of iterations in the algorithm.}
Algorithm \ref{algo:HyperMapper} shows the pseudo-code of the model-based search algorithm used in HyperMapper.
\begin{algorithm}
 \KwData{config. pool $X$, random sampling batch size $rs$}
 \KwResult{Pareto front $P$}
 $X_{out} \leftarrow$ sample $rs$ distinct configurations from $X$\;
 $Y_{ATE}, Y_{run} \leftarrow Evaluate(X_{out})$\;
 $M_{ATE} \leftarrow Fit\_Random\_Forest(X_{out}, Y_{ATE})$\;
 $M_{run} \leftarrow Fit\_Random\_Forest(X_{out}, Y_{run})$\;
 %$X_p \leftarrow X_p - X_{out}$ \;
 $P \leftarrow Predict\_Pareto(M_{ATE}, M_{run}, X)$\;
 %\While{$P \cap X_{out} \neq \oldemptyset$}{
 \While{$P - X_{out} \neq \oldemptyset$}{
 	$x \leftarrow P - X_{out}$\;
    %$x \leftarrow P \cap X_{out}$\;
    $y_{ATE}, y_{run} \leftarrow Evaluate(x)$\;
    $X_{out} \leftarrow X_{out} \cup x$\;
  	$Y_{ATE} \leftarrow Y_{ATE} \cup y_{ATE}$\;
  	$Y_{run} \leftarrow Y_{run} \cup y_{run}$\;
$M_{ATE} \leftarrow Fit\_Random\_Forest(X_{out}, Y_{ATE})$\;
    $M_{run} \leftarrow Fit\_Random\_Forest(X_{out}, Y_{run})$\;
        %$X_p \leftarrow X_p - x$\;
    $P \leftarrow Predict\_Pareto(M_{ATE}, M_{run}, X)$\;
 }
 \Return{$P$}\;
 \caption{Pseudo-code for HyperMapper. $-$ denotes set difference and $\cup$ denotes set union.}
 \label{algo:HyperMapper}
\end{algorithm}

% \fixmeL{Should add here a paragraph referring to the lines of the pseudo-code and explaining them. Even if the pseudo-code is clear enough and self-explanatory I this should be done for the journal version.}
%############################################################## 5 RESULTS ################################

\section{Experiments}
\label{evaluation}

In this section we describe how we evaluated our design space exploration. 
We begin by providing a more detailed description of the target platforms (Section \ref{platform}). 
We then briefly summarize our key results (Section \ref{section:results:overall}), before providing more detail on the results of the generalization of HyperMapper in Section \ref{section:results:dse}. For completeness and ease of comparison we partially report the results from our previous work on HyperMapper and design space exploration for the KFusion benchmark \cite{Bodin_PACT_2016}.

\subsection{Platforms}\label{platform}

For  our experiments we use a Hardkernel ODROID-XU3 platform based on the Samsung Exynos 5422, 
an ASUS T200TA with an Intel Atom Z3795, 
and a desktop computer with an NVIDIA GTX 780 Ti GPU. 

The Exynos 5422 includes a Mali-T628-MP6 GPU alongside ARM's big.LITTLE heterogeneous multiprocessing solution, consisting of four Cortex-A15 
``big'' performance tuned out-of-order processors, and four Cortex-A7 ``LITTLE'' energy tuned in-order processors. 
The Mali-T628-MP6 GPU consists of two separate OpenCL devices: one with four cores and another with two.
In our experiments we only use the 4-core OpenCL which excludes partitioning tasks across multiple GPU devices. 
This is a potential avenue to explore in order to deliver even higher performance within a power budget. The ODROID-XU3 supports OpenCL 1.1 and the GNU gcc compiler version 4.8.2 is used.

The ASUS Transformer T200 tablet contains an 
Intel Atom Z3795 SoC, which includes a quad-core Intel Atom CPU running at up to 2.4 GHz. An Intel HD Graphics GPU is 
also present, containing 6 execution units and running at up to 778 MHz. We use the open source Beignet~\cite{beignet} OpenCL runtime
which supports version 1.2 of the OpenCL standard and was produced by Intel's Open Technology Center. The GNU gcc compiler used is 5.3.1.

The desktop machine we used is a 8-core Intel Ivy Bridge E5-1620 v2 CPU augmented with a high-end discrete NVIDIA GPU GTX 780 Ti. The CUDA toolkit version is 7.5.18 and OpenGL version is 1.4. The machine runs Ubuntu OS kernel 14.04.4 and the GNU gcc compiler version 4.8.4 is used. 

We run KFusion on the ODROID-XU3 and ASUS mobile platforms using OpenCL, and ElasticFusion on the NVIDIA desktop using CUDA. 

%\subsection{Overall Results}
\subsection{Outcome in a glance}
\label{section:results:overall}

We observe that the default KFusion configuration provides a frame-rate of 6 FPS on the ODROID-XU3 embedded system. 
Our design space exploration results
show significantly better frame-rates with comparable accuracy.
For example, a configuration exists 
in the real-time range (29.09 FPS) and with a similar accuracy ATE compared to the default configuration (4.47 cm). These results are consistent also on the ASUS machine.
In addition, the selected best configurations perform well across a wide range of 83 mobile platforms crowd-sourced. These platforms are running the Android version of SLAMBench configured using the Pareto front, with speedups ranging from 2 to 12 over default. 

We observe that the default ElasticFusion configuration provides a frame-rate of 45 FPS on the NVIDIA desktop machine. 
Our design space exploration results
show significantly better frame-rates and better accuracy.
For example, a configuration exists that speeds-up the runtime by 1.52 compared to default while also improving accuracy. Another configuration shows a 2-fold improvement in accuracy (2.69 cm) compared to the default configuration (5.58 cm) with a speedup of 1.25.

Active learning effectively and consistently pushes the Pareto front toward better solutions. 
Taking into account the domain layer of the stack unleashes unprecedented performance trade-offs compared to the more usual compiler optimizations.

\subsection{Exploration}
\label{section:results:dse}
\begin{figure}[tb] 
                  
        \subfloat[ODROID-XU3][ODROID-XU3]{\centering\label{fig:ALParetoA}\includegraphics[width=0.95\columnwidth]{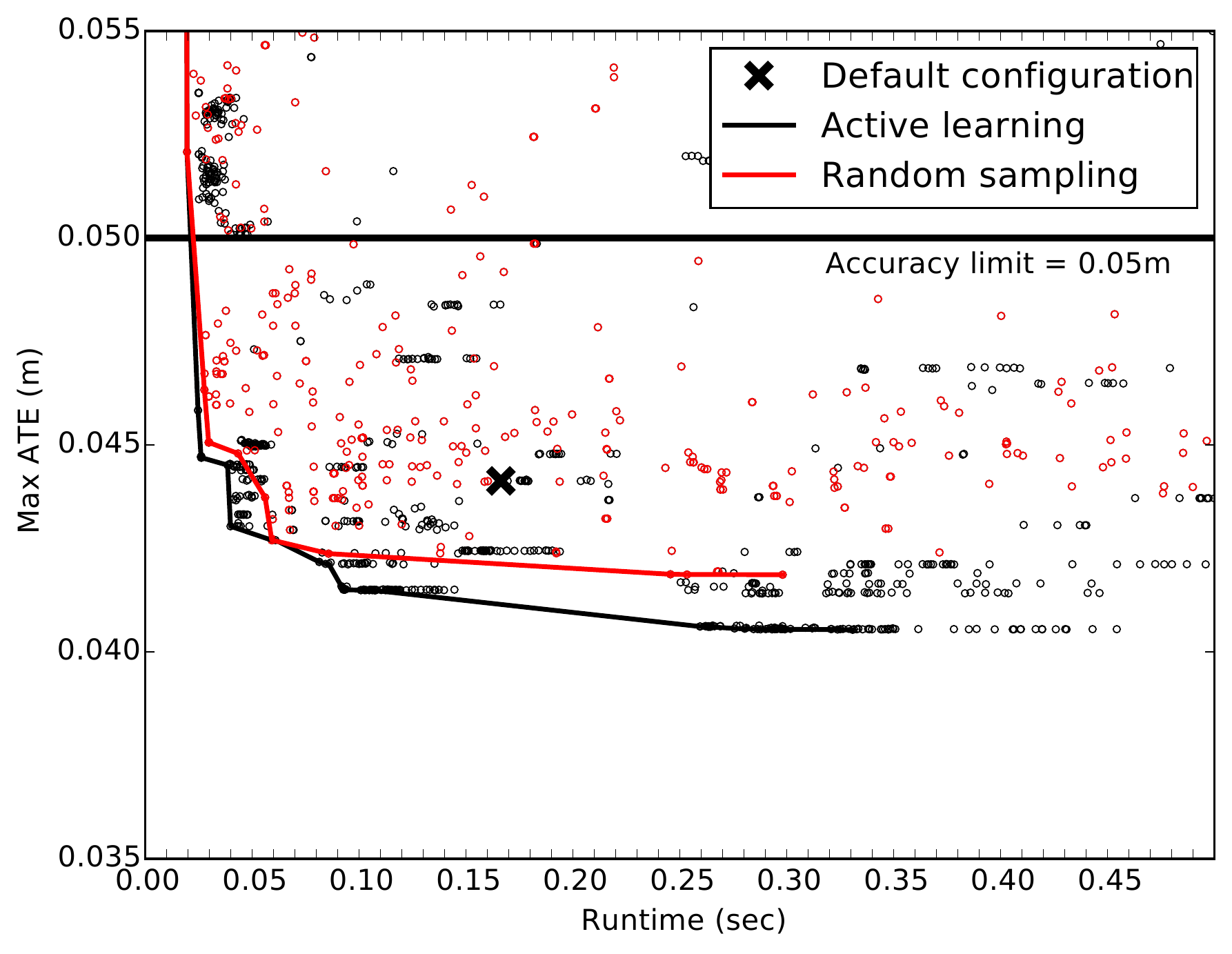}}\\
	\subfloat[ASUS T200TA][ASUS T200TA]{\centering\label{fig:ALParetoB}\includegraphics[width=0.95\columnwidth]{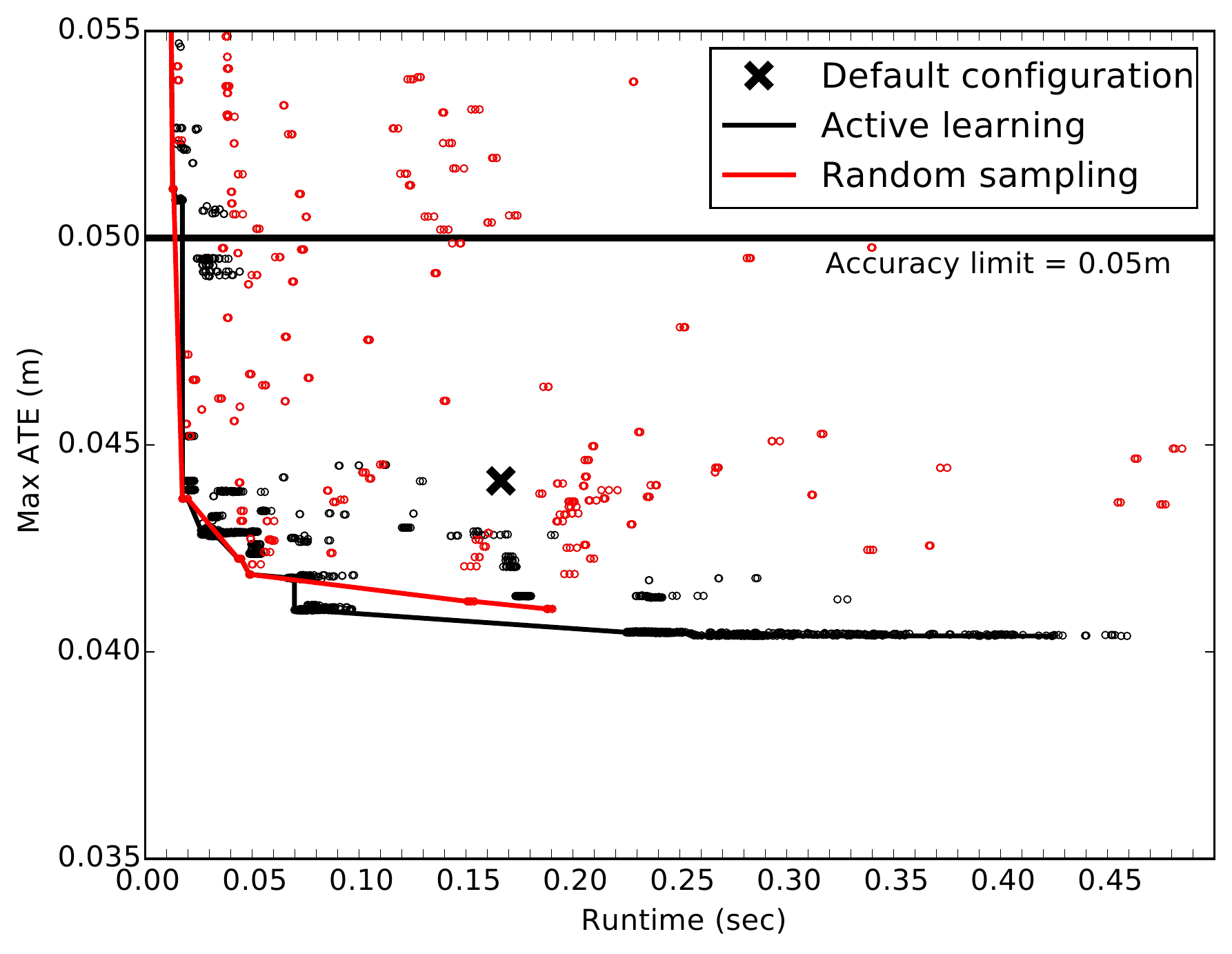}}\\
	\caption[Sampling solutions and active learning solutions]{
		Algorithmic design space exploration on the KFusion benchmark like shown in the original paper \cite{Bodin_PACT_2016}. Random sampling (red) and active learning (black).  
%\fixmeS{again psfrag is useful here}
	}
	\label{fig:ALPareto}
\end{figure}

The algorithmic space consists of application parameters described in Section \ref{section:methodology:KFusionDSE} and \ref{section:methodology:ElasticFusionDSE}. 
As described in Section \ref{section:machine learning}, we first sample this space at random, and 
then use active learning in order to push the Pareto front toward better solutions (refer to Figure \ref{fig:ALPareto}).

\paragraph{Sampling}
We draw 3,000 uniformly distributed random samples from the parameter space and 
evaluate the KFusion pipeline on the video stream; for both platforms the cumulated runtimes take roughly 5 days. 
By using random sampling, we observe that the Pareto front cannot be improved beyond 2,000 samples.
Thus, there is an inflection point beyond which random sampling is unproductive.

A similar number of uniformly distributed samples (2,400) is  used on ElasticFusion running on the NVIDIA machine.

\paragraph{Active learning}
In order to further explore optimal points in the design space, 
we employ active learning in conjunction with random decision forest 
(see \ref{section:machine learning}).
For the KFusion benchmark on ODROID-XU3 this produces 1,142 new samples after 6 iterations, thus increasing the total number of samples to 4,142.
Note that the number of samples produced per iteration is not constant as it depends on the predicted points' proximity to the Pareto front. 
We observe that the number of samples per iteration varies between 100 and 300.
The runtime of these new configurations was faster, close to a day, as most of these configurations were good configurations (accurate and fast).
The training of the random forest model was fast as well, less than two minutes for every iteration. 
With the ASUS T200TA platform, 1392 new points has been produced by active learning. On ElasticFusion on the NVIDIA platform 999 active learning points are collected. 

\paragraph{Active learning effectiveness} 
Figure~\ref{fig:ALPareto} shows the overall improvement of the Pareto front obtained with active learning (in black) compared to the Pareto obtained with random sampling (in red). 
For the ODROID-XU3 we observe that random sampling provides a set of 333 valid configurations, i.e. 333 configurations with a max ATE smaller than 5 cm. 
For the ASUS T200TA, we found 291 valid configurations during the sampling.
Furthermore, by using the active learning technique, we observe 642 new possible configurations with an ATE of less than 5 cm on the ODROID-XU3, and 665 on the ASUS T200TA. 
This means we have produced twice as many valid points as random sampling, for roughly a third of the number of samples.
These ratios are an indicator of the effectiveness of our active learning-based prediction model. 
There is a discrepancy between predicted and measured performance.
This is shown by the active learning points in Figure \ref{fig:ALPareto} that do not lie on the Pareto front.
Note that there are 36 points on the Pareto front for the ODROID-XU3 and 167 points for the ASUS T200TA.

\newcommand{\specialcell}[2][c]{%
	\begin{tabular}[#1]{@{}c@{}}#2\end{tabular}}

\begin{figure}
\captionsetup[subfigure]{labelformat=empty}
\subfloat[NVIDIA780Ti][NVIDIA 780 Ti]{\centering\label{fig:ALParetoC}\includegraphics[width=1.03\columnwidth]{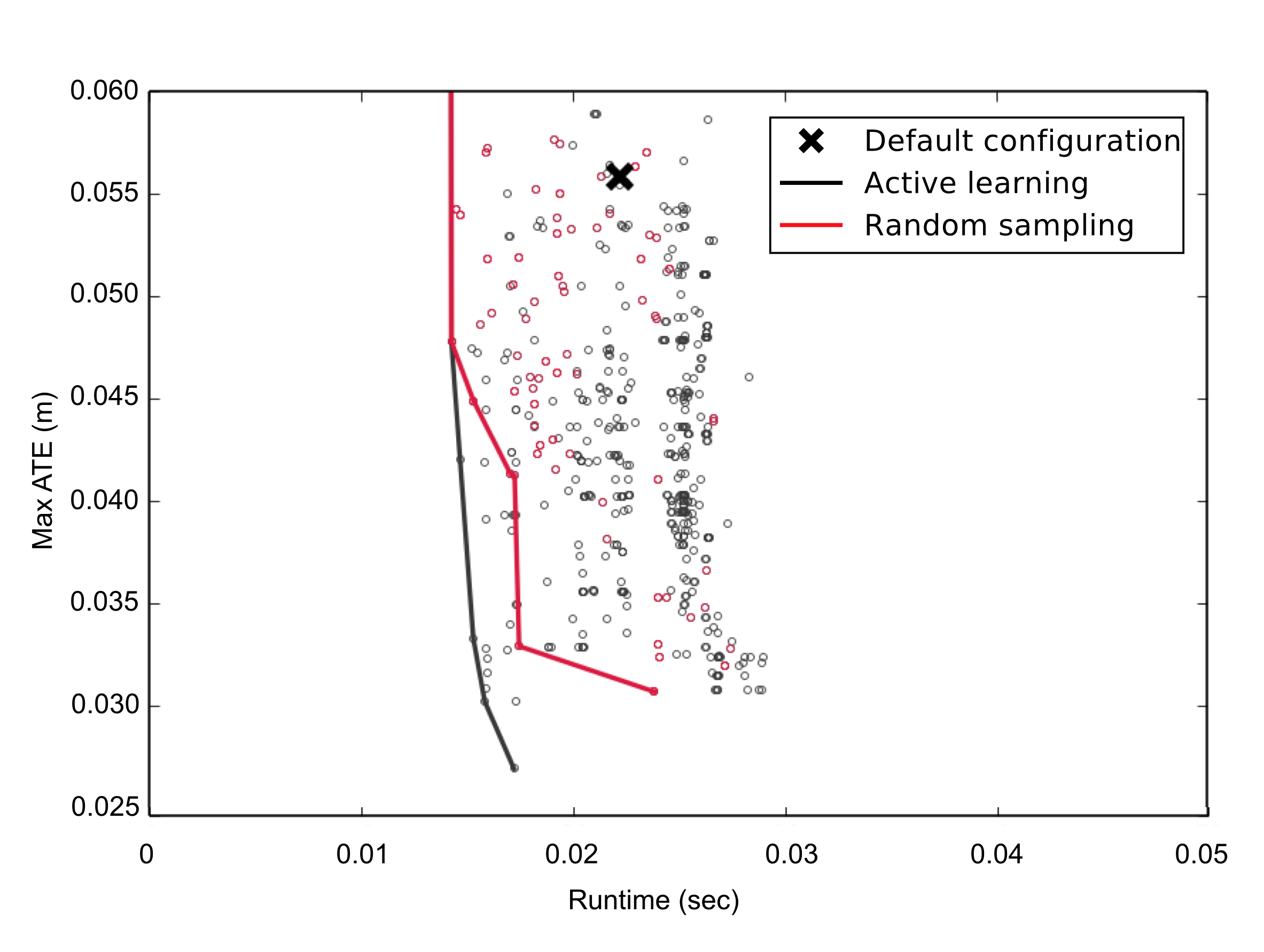}}
	\caption[Sampling solutions and active learning solutions]{
    Algorithmic design space exploration on the ElasticFusion benchmark. 
		Random sampling (red) and active learning (black).  
	}
	\label{fig:ALParetoElasticFusion}
\end{figure}

Figure~\ref{fig:ALParetoElasticFusion} shows the overall improvement on the ElasticFusion benchmark. 
This plot shows that HyperMapper is able to generalize the results obtained on KFusion on a fundamentally different and complex benchmark. Here again the active learning is consistently achieving an improvement in execution time and accuracy with respect to random sampling. See Table \ref{fig:pareto-icl-table} for details on the Pareto front and its algorithmic configuration values.

\begin{table*}                         \centering                                                                                                                        
\begin{tabular}{| M{0.095\textwidth} | M{0.065\textwidth} | M{0.08\textwidth} | M{0.035\textwidth}                                 
M{0.035\textwidth} M{0.055\textwidth} M{0.035\textwidth} M{0.085\textwidth}                                                         
M{0.055\textwidth} M{0.075\textwidth} M{0.07\textwidth} |}                                                                          
  \hline                                                                           \hline                                                                                                                            
  &  Error (m) & Runtime (s) & ICP & Depth & Confidence & SO3 & Close-Loops & Reloc & Fast-Odom & FTF RGB \\                                
 \hline\hline                                                                                                                       
  Default & 0.0558 & 22.2 & 10 & 3 & 10 & 1 & 0 & 1 & 0 & 0 \\                                                                      
 \hline                                                                                                                             
  Best speed        & 0.0420 & \textbf{14.6} & 5  & 6 &  9  & 0 & 0 & 1 & 1 & 0 \\                                                  
          & 0.0332 & 15.2 & 4  & 6 &  9  & 0 & 0 & 1 & 1 & 0 \\                                                                     
          & 0.0302 & 15.8 & 2  &10 &  4 & 0 & 0 & 1 & 1 & 0 \\                                                                      
  Best accuracy        &\textbf{0.0269} & 17.2 & 1 & 10 & 4 & 0 & 0 & 1 & 1 & 0 \\                                                  
 \hline                                                                                                                             
 \end{tabular}\caption{The Pareto efficiency points as a result of the design                                                       
 space exploration on the ICL NUIM Living Room 2 Dataset. On the top row are                                                        
 reported the results for the default configuration, highlighted in                                                                 
 boldface are the fastest and the most accurate configuration. }  
 
\label{fig:pareto-icl-table}                             
 \end{table*}

By using the described techniques to explore the algorithmic spaces, we have obtained a 6.35x improvement in execution time (best speed), compared to the default configuration on the ODROID-XU3 board. This important speedup can be explained by the fact that the application was tuned on a fundamentally different machine by the original developer, i.e. a NVIDIA Quadro GPU-based desktop. It is then not surprising that this default configuration performs poorly on a new target, the ODROID-XU3 or the ASUS in this case. 

The best speed up on the NVIDIA GPU is 1.52 while at the same time improving accuracy by 1.33, see Table \ref{fig:pareto-icl-table}. With respect to KFusion on ODROID-XU3 and ASUS, ElasticFusion on the NVIDIA GPU is a different test case. The ElasticFusion developers used a similar NVIDIA GTX machine to develop the application and, in addition, they used a brute force grid search to tune the parameters. HyperMapper is able to beat the human when compared to a similar setting than the hand-tuning one. Additionally HyperMapper is also able to find a configuration that performs 2x better in terms of accuracy.  

In \cite{Bodin_PACT_2016} we showed the  
correlation of the feature space with the runtime and the error metrics. We invite the reader to refer to that paper for correlation analysis.

\subsection{Crowd-sourcing}
\label{section:results:crowd-sourcing}
\begin{figure}
 	\centering
        \includegraphics[width=0.9\linewidth]{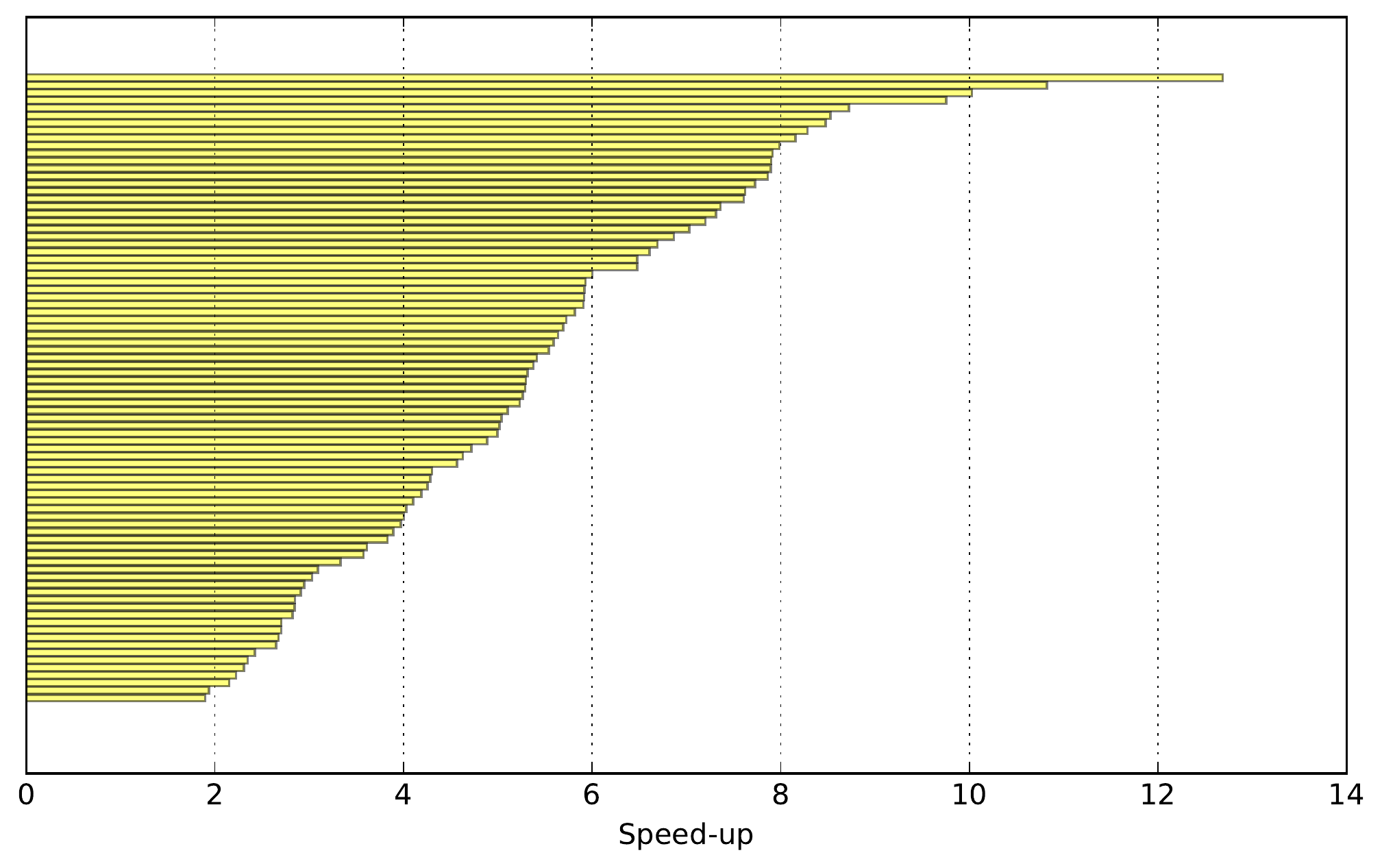}
 	\caption[KFusion Mobile]{The OpenCL KinectFusion has been run on 83 smart-phones and tablets from the market. For each device, we computed the speedup of the best configuration we found for the ODROID-XU3 and the original default configuration.} 
 	\label{fig:kfusion_mobile}
 \end{figure}

An Android app has been developed for the KFusion benchmark of SLAMBench \cite{SLAMBenchApp}. People can freely download the SLAMBench app and automatically run the 
%four 
best performing (best runtime) algorithmic configuration from the Pareto front computed on the ODROID-XU3 board together with the default configuration in order to benchmark their devices. For practical reasons, only 100 frames are run.
%from the trajectory 2 of the ICL-NUIM dataset. 
The app automatically collects the results and send it over the network to a centralized database for analysis purposes. In total 83 platforms ran the app. Figure \ref{fig:kfusion_mobile} shows this crowd-sourcing experiment. 
The plot shows the speedup over the default configuration run on the same device. 
%, of the best performing of the four configurations run. 
The speedup ranges between 2 and more than 12. This result confirms the hypothesis that best performing configurations found on one machine usually perform well also on different but similar machines. Specifically, most of the mobile devices in the market are ARM-based devices and these are the devices that populate our crowd-sourcing experiment. In \cite{roy2016exploiting} the authors show that there is a strong Pearson and Spearman correlation between configurations that perform well on one machine and configurations that perform well on another machine. And so that results on one machine can be often used to speed up a second machine. This is a form of zero-shot transfer learning that does not guarantee optimality but is showing to be effective. In \cite{roy2016exploiting} the authors also show that the zero-shot learning approach does not seem to work in general when the machines are fundamentally different, like for example from an Intel SandyBridge to an AppliedMicro X-Gene ARM 64-bit.
 
%\fixmeL{Cite some of the machines, the ones that perform well?
%It is only algorithmic? But some of the compiler transformations have been included: vectorisation (maybe thread coarsening and so on). 
%What happens at the best performing platforms? 
%What can we say about the different algorithmic points that score best? We can provide a plot with different colours 1 for each algo configuration.
%}

%############################################################## 6 RELATED WORK ################################
\section{Related work}
\label{section:RelatedWork}

The computer vision community primarily focuses on developing accurate algorithms~\cite{2014Handa, 2012Sturm}, 
almost always running on high-performance and power hungry systems. 
As computer vision technology becomes mature, a few benchmarks~\cite{2009Sdvbs, 2011Mevbench,7298655} 
have attempted to refocus research on runtime constrained contexts. 
Similarly, new challenges such as the Low-Power Image Recognition Challenge (LPIRC 2016) 
are emphasizing the importance of low-power embedded implementations of computer vision applications. 
In this context, recently SLAMBench~\cite{2015PAMELASLAMBench} enabled quantitative, 
comparable, and validatable experimental research in the form of a benchmark framework 
for dense 3D scene understanding on a wide range of devices. 
Adding energy consumption as a metric when evaluating computer vision applications, 
has enabled energy constrained systems such as battery-powered robots and embedded devices to become evaluation platforms. 
Zeeshan et al.~\cite{2016ICRA} is a first attempt at exploring SLAM configuration parameters trading off performance for accuracy on embedded systems. In \cite{bodin2016diplomat} the authors exploit the SLAMBench framework to explore optimization of multi-kernel application using a dataflow model. 

During the last two decades, several design space exploration techniques and frameworks 
have been used in a variety of different contexts ranging from embedded devices, to compiler research, and system integration.
%~\cite{2003Neema,1193228}. 
% Kang et al.~\cite{2011Kang} proposed a system which reduces the size of the design space by considering sets of design points to be equivalent. 
% Hu et al.~\cite{781351} present a user-guided design space exploration framework, 
% allowing the user to identify both good (and bad) design regions, and hence guide the subsequent search. 
Ansel et al.~\cite{ansel2014opentuner} introduced an extensible and portable framework for empirical performance tuning.
It runs an ensemble of search techniques systematically allocating larger budgets to those who perform well, 
using a multi-armed bandit optimal budget allocation strategy.  
Norbert et al. tackle the software configurability problem for binary~\cite{siegmund2012predicting} and 
for both binary and numeric options~\cite{siegmund2015performance} using a performance-influence model which is based on linear regression. 
They optimize for execution time on several examples exploring algorithmic and compiler spaces in isolation. 

In particular, machine learning (ML) techniques have been recently employed in both architectural and compiler research. 
Khan et al.~\cite{Khan:2007:UPC:1299042.1299059} employed predictive modeling for cross-program design space exploration in multi-core systems. 
The techniques developed managed to explore a large design space of chip-multiprocessors running parallel applications with low prediction error. 
In \cite{balaprakash2016automomml} Balaprakash et al. introduce AutoMOMML, an end-to-end, ML-based framework to build predictive models for objectives such as performance, and power. 
%Similarly, Ipek et al.~\cite{Ipek:2006:EEA:1168919.1168882} employed an artificial neural network to predict the impact on the performance of hardware parameters, e.g. cache sizes, buffer sizes, of a particular architecture. 
% Furthermore, Lee et al.~\cite{Lee:2006:AER:1168919.1168881} used polynomial regression to predict power and performance on a multiprocessor design space.  
%Add this reference in the camera ready
% Chen et al.~\cite{Chen:2014:ARA:2678373.2665688} suggest that a ML model can be used to produce a relative ranking of design points, rather than predicting their performance precisely.
% Camera ready: address the comments from Stefan
\cite{balaprakash2013active} presents the ab-dynaTree active learning parallel algorithm that builds surrogate performance models for scientific kernels and workloads on single-core, multicore and multinode architectures. 
In \cite{zuluaga2013active} the authors propose the Pareto  Active  Learning
(PAL) algorithm which intelligently samples the design space to predict the Pareto-optimal set.

%Furthermore, to the best of our knowledge, 
%we show that random forest in conjunction with active learning is effective 
%to focus the search for Pareto optimal configurations in this context.  

%################################################################ 7 CONCLUSION ##############################

\section{Conclusions and Future Work}
\label{section:conclusion}
3D scene understanding algorithms are generally complex and depend on a number  
of parameters that subtly interact with each other. Further configuration           
choices at the compiler and hardware level increase the mapping complexity,         
effectively making the manual tuning practice very difficult and often simply   
unfeasible. In this paper we demonstrated how the HyperMapper tool introduced in   
\cite{Bodin_PACT_2016} is effective across different SLAM algorithm                
implementations and different hardware platforms. Crucially, ElasticFusion's   
computational profile is very different from KFusion, confirming the robustness of our approach. %As with                            
%KFusion, we are able to attain XXXXX improvement over the baseline                  
%configuration hand-tuned by domain experts, in all the considered performance   
%parameters.                                         %                                
%We also quantified the gap between the results achievable in our controlled         
%experimental environment and real world devices. 
The crowd-sourced data allowed us to access a variety of devices    
and simulation settings, e.g. HW platforms, compiler and operating system versions, and showed consistent and important speedups on today market mobile platforms.  
%This has proven to be very fruitful in terms of insights on the real world.
                                                                                    
In future work,  we aim to add more SLAM input data-sets in order                   
to encompass a larger number of scenarios, providing more            
breadth in terms of trajectories and real-world use cases. Regarding                
HyperMapper, we will not only investigate new techniques to reduce the dimension of the design space, but also more advanced transfer learning and resampling techniques. A powerful               
application of such techniques would be to treat multiple algorithms,               
compilers and platforms on the same tuning session, effectively enacting            
an algorithm selection tailored to the specific operative scenario.
In this case domain-specific languages (DSLs) \cite{2009Nardi,2012Nardi} would be the perfect vehicle to harness the algorithmic exploration automatically. 
\section{Acknowledgments}

We acknowledge funding by the EPSRC grant PAMELA
EP/K008730/1.
We thank the PAMELA Steering Group for the useful discussions.
We also thank the various students that contributed to this project, 
in particular Denise Carroll and Alfonso White.
\bibliographystyle{abbrv}

% that's all folks
\end{document}